\PassOptionsToPackage{table}{xcolor}
\documentclass[sigconf]{acmart}
\usepackage{float}
\usepackage{balance}
\AtBeginDocument{%
  }
\usepackage{multirow} 
\usepackage{tabularx}
\usepackage{arydshln}
\usepackage{subcaption}
\usepackage{enumitem}
\usepackage{booktabs}
\usepackage{bbm}
\usepackage{graphicx}
\definecolor{pastelblue}{RGB}{235, 243, 255} 
\definecolor{pastelred}{RGB}{255, 240, 240}

\copyrightyear{2026}
\acmYear{2026}
\setcopyright{cc}
\setcctype{by}
\acmConference[KDD '26]{Proceedings of the 32nd ACM SIGKDD Conference on Knowledge Discovery and Data Mining V.2}{August 09--13, 2026}{Jeju Island, Republic of Korea}
\acmBooktitle{Proceedings of the 32nd ACM SIGKDD Conference on Knowledge Discovery and Data Mining V.2 (KDD '26), August 09--13, 2026, Jeju Island, Republic of Korea}
\acmDOI{10.1145/3770855.3818500}
\acmISBN{979-8-4007-2259-2/2026/08}
\begin{document}

\title{UME: A Unified Meta-Generalization Framework for Cross-Domain ETA}

\author{Duo Wang}
\email{wang_d@stu.pku.edu.cn}
\affiliation{%
  \institution{Peking University}
  \city{Beijing}
  \country{China}
}

\author{Qiong Wu}
\email{wu_qiong@stu.pku.edu.cn}
\affiliation{%
  \institution{Peking University}
  \city{Beijing}
  \country{China}
}

\author{Jianguo Wu}
\authornote{Corresponding author.}
\email{j.wu@pku.edu.cn}
\affiliation{%
  \institution{Peking University}
  \city{Beijing}
  \country{China}
}

\author{Ruiyu Xu}
\email{xuruiyu@meituan.com}
\affiliation{%
  \institution{Meituan}
  \city{Beijing}
  \country{China}
}

\author{Jinhui Yi}
\email{yijinhui@meituan.com}
\affiliation{%
  \institution{Meituan}
  \city{Beijing}
  \country{China}
}
\author{Zhonggen Sun}
\email{sunzhonggen@meituan.com}
\affiliation{%
  \institution{Meituan}
  \city{Beijing}
  \country{China}
}
\author{Zhentao Zhang}
\email{zhangzhentao@meituan.com}
\affiliation{%
  \institution{Meituan}
  \city{Beijing}
  \country{China}
}
\author{Yu Zhang}
\email{zhangyu420@meituan.com}
\affiliation{%
  \institution{Meituan}
  \city{Beijing}
  \country{China}
}
\author{Ke Xing}
\email{xingke@meituan.com}
\affiliation{%
  \institution{Meituan}
  \city{Beijing}
  \country{China}
}
\author{Yongjun Yin}
\email{yinyongjun@meituan.com}
\affiliation{%
  \institution{Meituan}
  \city{Beijing}
  \country{China}
}
\author{Zishuo Li}
\email{lizishuo03@meituan.com}
\affiliation{%
  \institution{Meituan}
  \city{Beijing}
  \country{China}
}
\author{Jianwen Huang}
\email{huangjianwen04@meituan.com}
\affiliation{%
  \institution{Meituan}
  \city{Beijing}
  \country{China}
}

\renewcommand{\shortauthors}{Duo Wang et al.}

\begin{abstract}

Accurate Estimated Time of Arrival (ETA) prediction on checkout page is crucial in instant logistics for enhancing user satisfaction, optimizing dispatching, and controlling operational costs. In international on-demand delivery platforms, where ETA data originates from diverse countries or regions with different patterns, multi-domain modeling is of great importance and has been widely adopted. However, existing methods still face three critical challenges in real-world deployment. First, current multi-domain models struggle to generalize to completely unseen domains, failing to achieve zero-shot prediction during the initial cold-start phase. Second, cross-domain feature spaces are often assumed to be consistent, whereas new domains commonly suffer from structural missingness of offline (statistical) features due to the lack of historical data. Third, such feature missingness often compels industrial systems to model mature and cold-start domains separately, hindering knowledge transfer and increasing maintenance overhead. To address these challenges, we propose \textbf{UME}, a \textbf{U}nified \textbf{M}eta-generalization framework for \textbf{E}TA, which enables zero-shot domain generalization on unseen domains while achieving superior performance on mature source domains. Specifically, UME integrates a unified dual-branch architecture with a novel meta-learning mechanism that employs a hypernetwork-based meta learner. By leveraging domain-level knowledge and instance-level context, the meta learner empowers three meta modules to dynamically modulate feature gating, expert attention, and final prediction, capturing cross-domain correlations and facilitating intra-domain adaptation. A knowledge distillation strategy is further introduce to enhance performance. UME has now been deployed in Meituan-keeta delivery platform (the largest international food delivery platform in China). Extensive offline experiments and online A/B tests demonstrate that UME significantly outperforms existing baselines.

\end{abstract}


\maketitle

\section{Introduction}
\label{sec:intro}

\begin{figure}[htbp]
    \vspace{-0.15cm}
    \centering  
    \includegraphics[width=\linewidth]{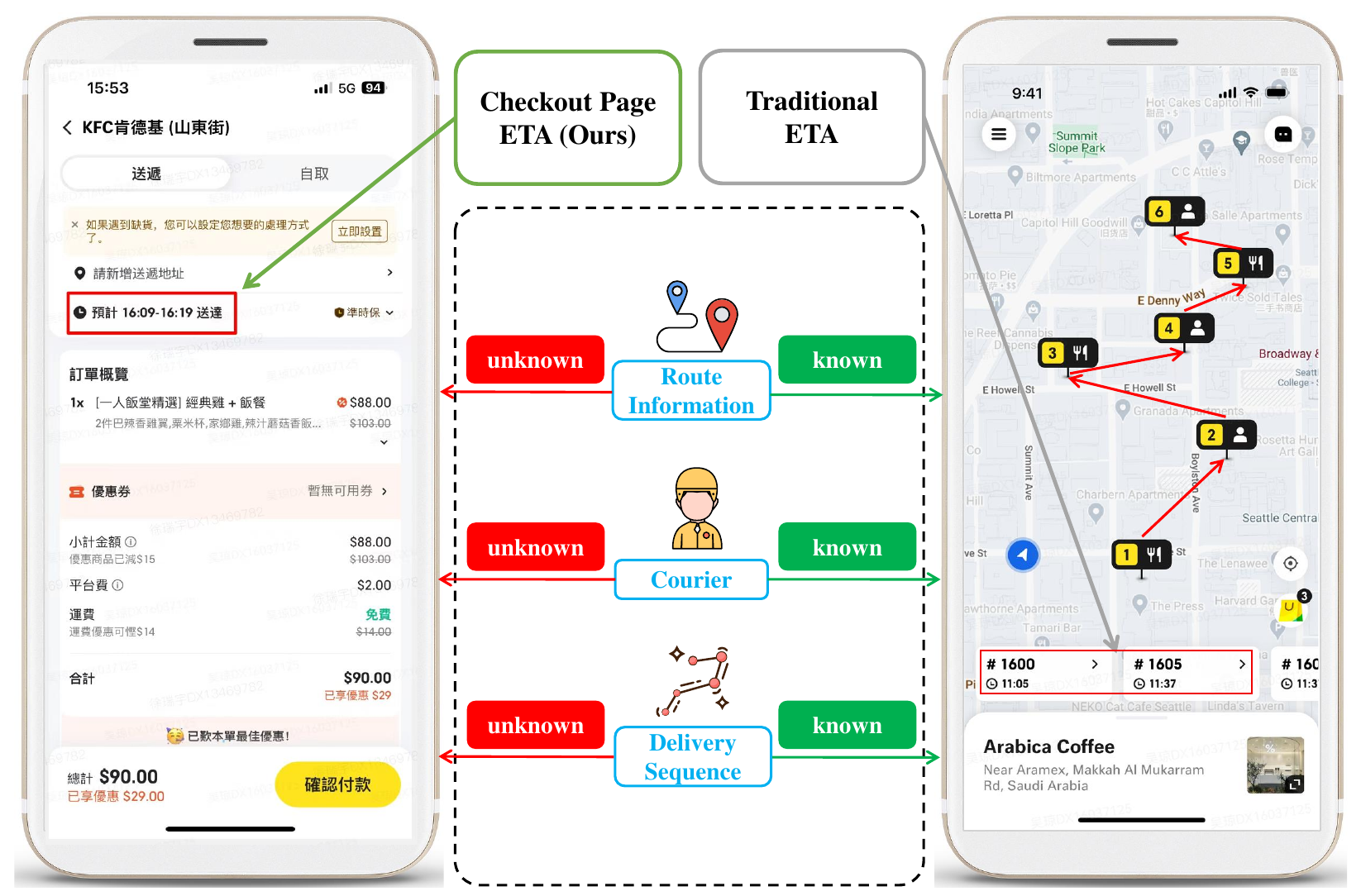} 
    \vspace{-0.55cm}
    \caption{Checkout Page ETA} 
    \label{fig:ETA}      
    \vspace{-7pt}
\end{figure}

With the rapid growth of instant logistics and on-demand services, user expectations for timely fulfillment have become increasingly stringent~\cite{yi2023deepsta,hildebrandt2022supervised, yi2024rccnet}. In this context, accurate Estimated Time of Arrival (ETA) provided at the checkout stage, referred to as Checkout Page ETA in Fig.~\ref{fig:ETA}, is of critical importance~\cite{zhu2020order}. High-precision ETA improves user satisfaction, supports intelligent dispatching, and reduces operational costs~\cite{wang2018learning}. Consequently, ETA modeling has become a central research focus in both academia and industry~\cite{michalopoulou2023meal}.

Although substantial progress has been made in ETA~\cite{arifi2024study}, most prior work focuses on route-based or origin–destination (OD)-based Travel Time Estimation (TTE). In contrast, Checkout Page ETA, as shown in Fig.~\ref{fig:ETA}, is shaped not only by traffic dynamics but also by complex business logic and multi-party behaviors, covering merchant preparation, system dispatching, and courier delivery~\cite{ruan2020doing}. Moreover, Existing methods typically assume a single homogeneous domain~\cite{jin2022selective, zhu2020order}, an assumption that rarely holds in real-world platforms. In internationally operated platforms, such as Keeta and DoorDash, data exhibits significant cross-country and cross-city heterogeneity in mobility patterns, urban infrastructure, dispatch strategies, and consumer behavior~\cite{zheng2014urban}. Such variations induce substantial regional shifts in ETA-related feature distributions. 


From a multi-domain perspective, ETA data from different countries or cities can be viewed as distinct domains with heterogeneous feature distributions. Training separate models for each domain captures such differences but suffers from data sparsity in small domains and high engineering overhead. Conversely, simply pooling all data without accounting for domain distinctions to train a single model may introduce cross-domain interference and degrade performance~\cite{tang2020progressive}. Multi-Domain Learning (MDL) methods offer a potential solution by jointly modeling shared and domain-specific patterns within a unified framework~\cite{zhou2023hinet, chen2020scenario}. 

\begin{figure}[htbp]
    \vspace{-0.25cm}
    \centering  
    \includegraphics[width=\linewidth]{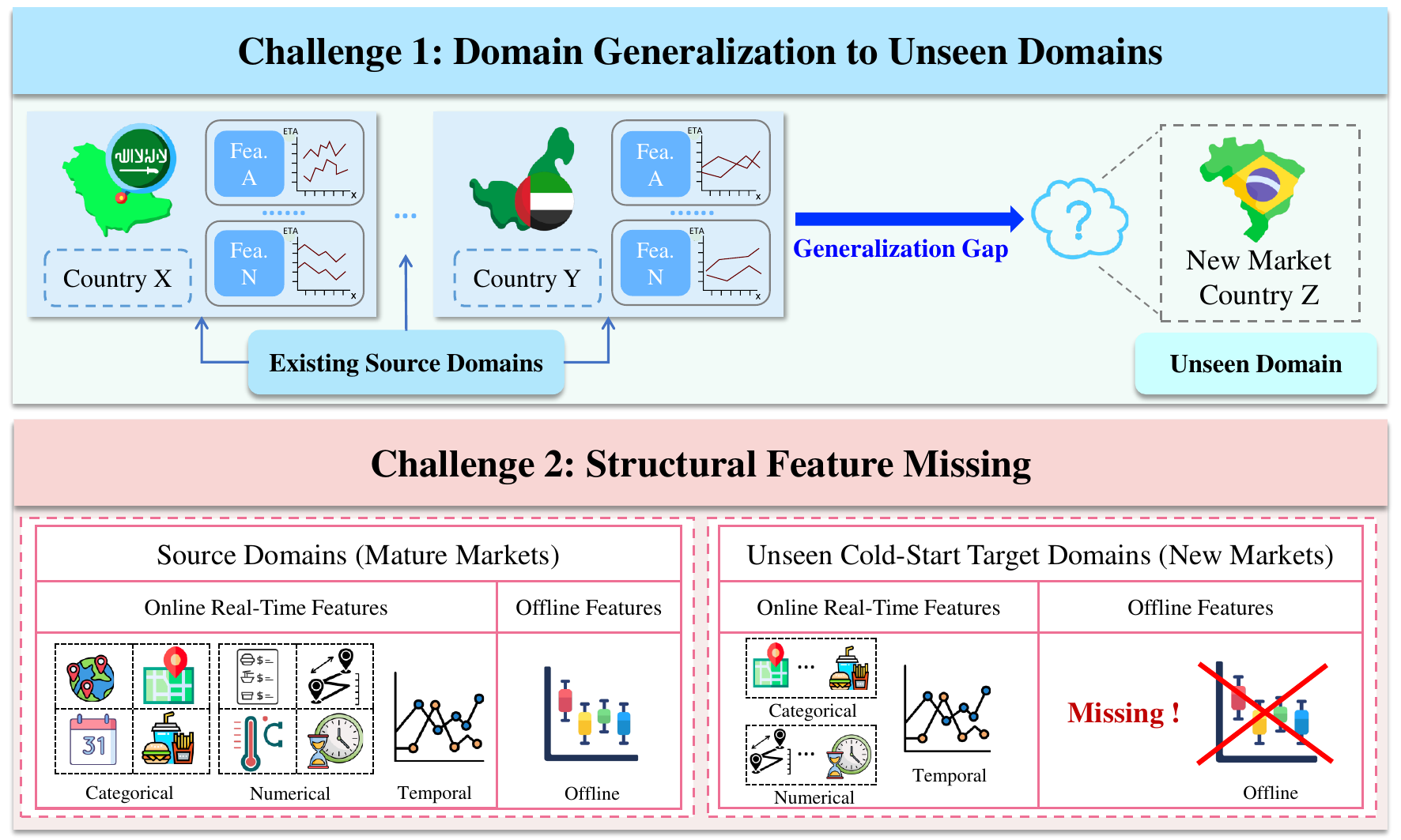} 
    \vspace{-0.55cm}
    \caption{Challenges in cross-domain ETA} 
    \label{fig:challenge}      
    \vspace{-15pt}
\end{figure}

However, existing MDL approaches still face three key limitations. First, most MDL methods struggle to generalize to completely unseen domains~\cite{wang2022generalizing}. As illustrated in Fig.~\ref{fig:challenge}, international platforms continuously expand into new cities or countries, where historical data are typically unavailable during the initial cold-start phase. Nevertheless, the ETA system must be operational from the very beginning of a market launch, requiring the model to rely solely on knowledge learned from existing domains to achieve zero-shot generalization. This setting fundamentally constitutes a Domain Generalization (DG) problem~\cite{wang2022generalizing}, in which the model is trained on a limited set of source domains (e.g., countries or cities with long-term operations) but must be deployed in unseen target domains at test time. Second, existing MDL and DG approaches typically assume consistent feature spaces across domains~\cite{pan2009survey,wang2020heterogeneous}. In practice, mature domains contain rich offline features (aggregated statistical features), whereas newly launched domains lack such historical features, resulting in structural feature missingness in addition to distribution shifts. Third, these limitations often lead to fragmented industrial solutions~\cite{cao2025industrial}. Specifically, complex MDL models are deployed for feature-complete mature domains, while simplified or rule-based models are used for feature-missing cold-start domains. This fragmented setup not only incurs high system maintenance and iteration costs~\cite{coop2021cost}, but also results in “knowledge discontinuity”, where knowledge accumulated in mature domains cannot be effectively transferred to new ones~\cite{wang2018cross}, and “feature dependency”, in which heavy reliance on offline features makes the system fragile under sudden abnormal conditions, such as severe weather~\cite{wani2025ten}. Therefore, building a general unified ETA framework that simultaneously supports feature-complete mature domains and feature-missing cold-start domains remains an open challenge. To the best of our knowledge, we are the first to identify and formalize this long-standing industrial challenge in ETA prediction: how to unify multi-domain learning and domain generalization under structural feature missingness.

To address these challenges, we propose \textbf{U}\textbf{M}\textbf{E}, a \textbf{U}nified \textbf{M}eta-generalization framework for \textbf{E}TA, which
enables zero-shot domain generalization on unseen target domains while
achieving superior performance on source domains. In UME, each source domain is equipped with an additional Generalization Branch trained solely on basic features to simulate feature-missing cold-start scenarios. In addition, a meta-learning mechanism is introduced to facilitate both cross-domain and intra-domain adaptation. The shared meta learner captures cross-domain generalization patterns from multiple source domains during training, enabling it to generate appropriate domain-specific parameters for unseen domains at test time, thereby achieving zero-shot domain generalization.

In summary, the contributions of this paper are as follows:

\begin{itemize}[leftmargin=*, noitemsep, topsep=2pt ]
    \item  We propose the first unified framework (UME) for ETA prediction that simultaneously addresses multi-domain modeling for source domains and domain generalization for unseen cold-start target domains. Notably, the proposed framework is generic and can be readily extended to application fields beyond ETA.
    \item  We develop a unified dual-branch architecture integrated with a novel meta-learning mechanism that employs a hypernetwork-based meta learner. By leveraging domain-level knowledge and instance-level context, the meta learner dynamically modulates feature gating, expert attention, and final prediction through three meta modules, capturing cross-domain correlations and facilitating intra-domain adaptation. Furthermore, a knowledge distillation strategy is introduced to bridge the information gap caused by structural feature missingness.
    
    \item  UME has now been deployed in Meituan-keeta delivery platform (the largest international food delivery platform in China), and extensive offline experiments and online A/B tests demonstrate that UME achieves state-of-the-art performance.
\end{itemize}

\section{Related Works}

\subsection{Multi-Domain Learning}

Multi-domain learning aims to capture shared and domain-specific patterns across domains using a unified model~\cite{joshi2012multi,yang2014unified}, and is often referred to as Multi-Scenario Learning (MSL) in industry. To handle domain discrepancies, HMoE~\cite{li2020improving} adopts a hierarchical Mixture-of-Experts architectur. STAR~\cite{sheng2021one} introduces a star-topology structure to adjust domain-specific networks, and M2M~\cite{zhang2022leaving} and 3MN~\cite{zhang20233mn} leverage meta-learning to generate scenario-specific parameters. More recently, MultiLoRA~\cite{song2024multilora} proposes a multi-directional LoRA to address cross-domain shifts. UFIN~\cite{tian2025ufin} further incorporates LLMs to learn universal semantic features across diverse domains. Despite these advances, existing methods typically rely on fixed and known domain labels and assume consistent feature spaces, failing to effectively transfer knowledge from mature source domains to feature-missing unseen cold-start domains.



\subsection{Domain Generalization}

Domain Generalization (DG) aims to train models on multiple source domains to generalize to unseen target domains~\cite{zhou2022domain}. Early approaches focused on learning domain-invariant representations~\cite{muandet2013domain}. Beyond simple invariance learning, IRM~\cite{arjovsky2019invariant} seeks representations that yield an optimal classifier across all training environments, while ARM~\cite{zhang2021adaptive} adopts a meta-learning paradigm to improve adaptation to concept shifts using unlabeled data. Qu et al.~\cite{qu2022hmoe} further proposed a hypernetwork-based MoE that dynamically generates expert weights conditioned on inputs. Despite these advances, these methods become inapplicable when unseen target domains suffer from feature missingness. Although SimMMDG~\cite{dong2023simmmdg} improves DG robustness under modality missingness by inferring missing modalities, it is difficult to apply in our scenario, where an entire set of high-value offline features is structurally absent due to the lack of historical data accumulation.

\subsection{ETA}

Estimated Time of Arrival (ETA) traditionally refers to travel time between an origin and destination~\cite{zhang2023delivery, wang2018learning}. Existing methods are generally categorized as route-based~\cite{chen2022interpreting, sui2024congestion}, which aggregate segment-level travel times along a planned route, and origin–destination (OD)-based approaches~\cite{han2023ieta, hu2020stochastic, yi2024learning}, which estimate travel time directly from the OD pair and departure time. In this work, we focus on Checkout Page ETA, also known as Order Fulfillment Cycle Time (OFCT)~\cite{zhu2020order, wei2024process}. Unlike traditional ETA, OFCT must be predicted before order creation, when key information—such as courier assignment and delivery route—is unavailable. Moreover, OFCT is affected by additional uncertainties, including merchant preparation time and dispatch delays. These characteristics make conventional methods inadequate and require a framework capable of modeling broader uncertainty.

\section{Problem Formulation}
\label{sec:Problem Formulation}

\begin{figure*}[tbp]
    \centering  
    \includegraphics[width=1.0\textwidth]{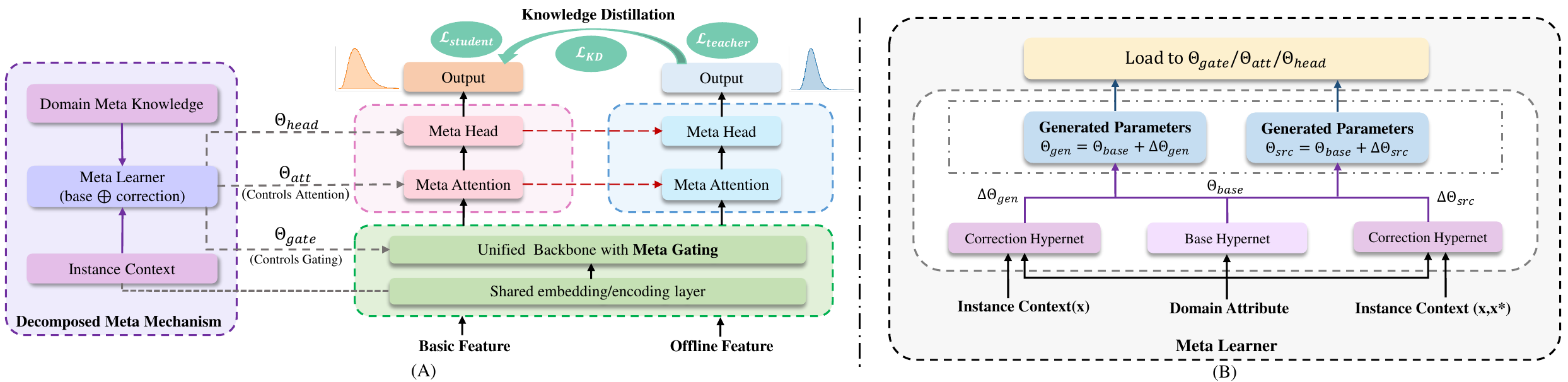} 
    \vspace{-0.55cm}
    \caption{\small Overview of UME, illustrating the Unified Dual-Branch Network empowered by a novel meta-learning mechanism.} 
    \label{fig:overview}      
    \vspace{-8pt}
\end{figure*}

Let $\mathcal{X}$ denote the basic online feature space, comprising real-time numerical features, categorical features (e.g., order AOI) and temporal features, denoted as $x \in \mathcal{X}$. Specifically, numerical features record variables such as delivery distance and the number of idle riders. Temporal features are time series that record system state sequences constructed over sliding time windows (e.g., the number of order assignments in an area during the past 10 minutes). Features in this space are instantly available and can be acquired in real-time. Let $\mathcal{X}^*$ denote the offline feature space, composed of statistical features derived from long-term historical accumulation, such as the merchant's historical average meal preparation time and the region's historical traffic speed distribution, denoted as $x^* \in \mathcal{X}^*$. Constrained by data accumulation, these features are available only in mature source domains and are structurally missing in cold-start domains. Let $\mathcal{Y}$ denote the target prediction space (i.e., arrival time), represented as $y \in \mathcal{Y}$.

Assume we have $M$ mature source domains, denoted as $\mathcal{D}_{train} = \{\mathcal{S}^m\}_{m=1}^M$, each following distribution $P_{XX^*Y}^m$ with feature-complete samples
\begin{equation}\mathcal{S}^m = \{(x_i^m, x_i^{*m}, y_i^m)\}_{i=1}^{N_m},\end{equation}where $N_m$ denotes the number of samples in the $m$-th domain. Consequently, the model can be trained utilizing both $x$ and $x^*$.

Consider a potential set of unseen target domains (e.g., new cities scheduled for launch), denoted as $\mathbb{T} = \{\mathcal{T}^1, \mathcal{T}^2, \dots \}$. For any $\mathcal{T}\in\mathbb{T}$, the data distribution $P_{XY}^{\mathcal{T}}$ differs from source domains, and offline features $x^*$ are structurally missing. Hence, the unseen target samples are\begin{equation}\mathcal{T} = \{(x_j^t, y_j^t)\}_{j=1}^{N_t},\end{equation}and predictions rely solely on $\mathcal{X}$.

Our objective is to develop a unified model $\mathcal{F}$ that supports both unseen domain generalization and multi-domain modeling. Specifically, for any unseen target domains $\mathcal{T}\in\mathbb{T}$, $\mathcal{F}$ predicts using only basic features, while for source domains $\mathcal{S}^m$, it leverage the offline feature $x^*$ to achieve superior performance.

\section{Methodology}

\subsection{The proposed UME framework}




The overview of UME is illustrated in Fig.~\ref{fig:overview}. The framework consists of three key components: a Unified Dual-Branch Network, a novel Meta-Learning Mechanism and a Distillation Strategy. Specifically, to reduce model maintenance costs and enable mutual enhancement between cold-start and mature-operation prediction tasks, we avoid the simple ensemble of two independent models. Instead, we simulate the feature-missing condition of target domains during training on source domains. Based on this idea, we design a novel Unified Dual-Branch Network (UDBN). This architecture constructs two parallel inference pathways within a single model: a Source Branch, which leverages full features $(X, X^*)$, and a Generalization Branch, which relies solely on basic features $X$. The former is tailored for prediction in mature domains, while the latter is designed for unseen cold-start target domains. This design yields a unified framework that supports the full lifecycle of ETA prediction. 

Furthermore, to equip the model with multi-domain modeling and domain generalization capability, we introduce a hypernetwork-based meta mechanism. Rather than directly optimizing fixed model parameters, this mechanism learns meta-knowledge about how to dynamically generate specific target parameters of UDBN conditioned on domains. In the following sections, we elaborate on these designs.

\subsubsection{Unified Dual-Branch Network}


In this section, we introduce UDBN, which serve as the feature extraction backbone of our framework. Although the two branches share the same backbone architecture, they operate under an information asymmetric setting.





\paragraph{Feature Encoding and Embedding}

\begin{figure}[htbp]
    \vspace{-0.2cm}
    \centering  
    \includegraphics[width=\linewidth]{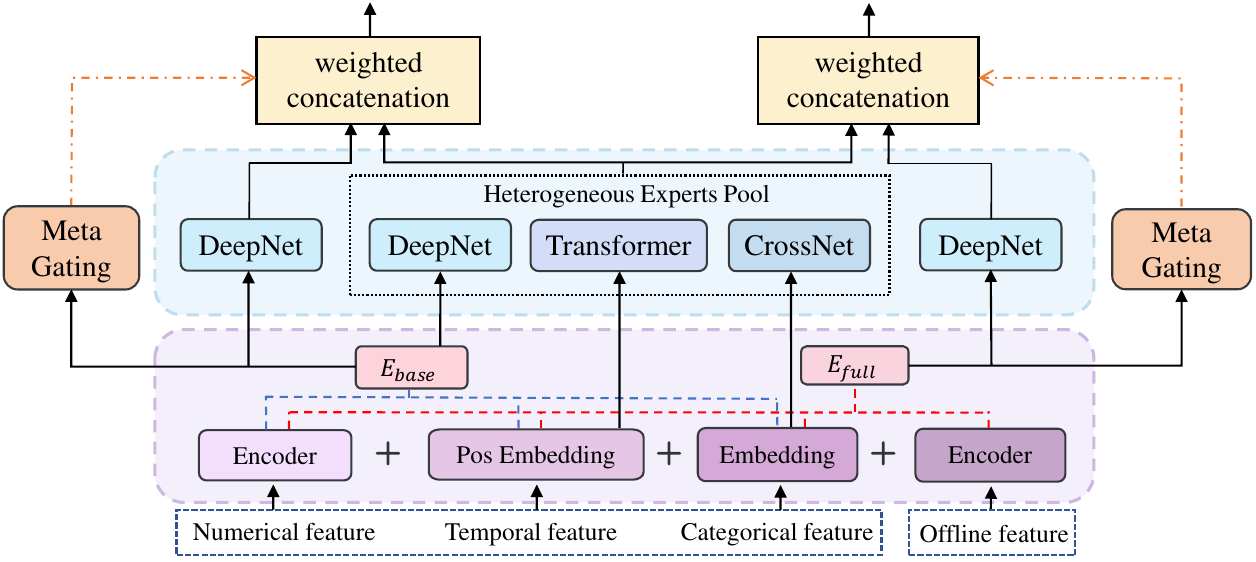} 
    \vspace{-0.54cm}
    \caption{\small Unified Backbone with Meta Gating} 
    \label{fig:UAM}      
    \vspace{-10pt}
\end{figure}

As illustrated in Fig.~\ref{fig:UAM}, the inputs are first processed by a unified bottom module, where the feature space is decomposed into four subspaces as mentioned in Section~\ref{sec:Problem Formulation}: real-time numerical features $\mathbf{x}_{num}$, categorical features $\mathbf{x}_{cat}$, temporal features $\mathbf{x}_{tmp}$, and offline features $\mathbf{x}_{off}$. Parallel encoders and preprocessing modules are applied to each subspace. Real-time numerical features are encoded via an MLP:
$\mathbf{e}_{num} = \phi_{mlp}(\mathbf{x}_{num}) \in \mathbb{R}^{d_{num}}$.
Categorical features are mapped through embedding layers:
$\mathbf{e}_{cat} = \text{Embed}(\mathbf{x}_{cat}) \in \mathbb{R}^{d_{cat}}$.
Temporal features are processed using an aggregation layer with local temporal pooling and positional embeddings:
$\mathbf{e}_{tmp} = \text{Agg}(\mathbf{x}_{tmp}) + \mathbf{p}_{pos} \in \mathbb{R}^{d_{tmp}}$.
Offline features are encoded by another independent MLP:
$\mathbf{e}_{off} = \phi_{off}(\mathbf{x}_{off}) \in \mathbb{R}^{d_{off}}$. Consequently, we obtain the joint representation of basic features $\mathbf{E}_{base} = \text{Concat}(\mathbf{e}_{num},\allowbreak \mathbf{e}_{cat}, \text{AverPool}(\mathbf{e}_{tmp}))$ and the joint representation of full features $\mathbf{E}_{full} = \text{Concat}(\mathbf{E}_{base}, \mathbf{e}_{off})$.



Based on the encoded features, we introduce a Heterogeneous Multi-View Experts layer, where heterogeneous experts process distinct feature subspaces from multiple complementary “views”. We define a general shared expert pool $\mathcal{E}_{shared}$ consisting of $K$ view experts, along with two branch-specific private experts $\mathcal{E}_{private}$ to capture branch-specific information. The Shared View Experts are shared between the Generalization Branch and the Source Branch and exclusively utilize basic feature subspaces to extract multi-view common features.

Specifically, for ETA data, we instantiate three shared view experts as shown in Fig.~\ref{fig:UAM}. The Deep View Expert ($E_{sh}^{dnn}$) models holistic interactions over the joint basic representation $E_{base}$ via a DNN: $h^{deep}_{sh} = E_{sh}^{dnn}(E_{base})$. The Cross View Expert ($E_{sh}^{cross}$) captures high-order combinational interaction effects of categorical features using a Cross Network~\cite{wang2021dcn} on $e_{cat}$: $h^{cross}_{sh} = E_{sh}^{cross}(e_{cat})$. The Temporal View Expert ($E_{sh}^{trans}$) leverages Transformer-based self-attention over $e_{tmp}$ to capture long-range temporal dynamics, with the last hidden state taken as the representation: $h^{tmp}_{sh} = E_{sh}^{trans}(e_{tmp})$. For branch-specific private experts, the Generalization View Expert ($E_{gen}$) takes the basic representation $E_{base}$ as input and serves the generalization branch. Its objective is to encourage the extraction of more informative representations from $E_{base}$ in the absence of offline features, complementing the shared views: $
h_{gen} = E_{gen}(E_{base}).$ The Source View Expert ($E_{src}$) takes the full representation $E_{full}$ as input and serves the Source Branch, fully leveraging historical statistical features in mature source domains: $h_{src} = E_{src}(E_{full})$. Both branch-specific experts are instantiated using the DNN architecture illustrated in Fig.~\ref{fig:UAM}. This design explicitly addresses feature asymmetry across branches and enables unified modeling under both feature-complete and feature-missing settings.


Next, we introduce the two branches empowered by the meta-learning mechanism.

\subsubsection{Meta learning Mechanism}

To achieve multi-domain modeling and cross-domain generalization, we introduce a noval meta-learning mechanism in which a hypernetwork-based meta learner is designed to learn explicit inter-domain correlations and empowers three meta modules to adaptively modulate UDBN in a domain-specific manner. Distinct from conventional parameter-generating meta-learning approaches~\cite{zhang2022leaving,pan2019urban}, our meta learner is built upon a decomposed meta mechanism. By conditioning on both domain-level knowledge and instance-level context, it enables more precise and adaptive parameter generation. Further details about the meta learner are provided in Section~\ref{meta learner}. We next introduce the three meta modules empowered by the meta learner.

\begin{figure}[htbp]
\vspace{-0.2cm}
    \centering  
    
    \includegraphics[width=\linewidth]{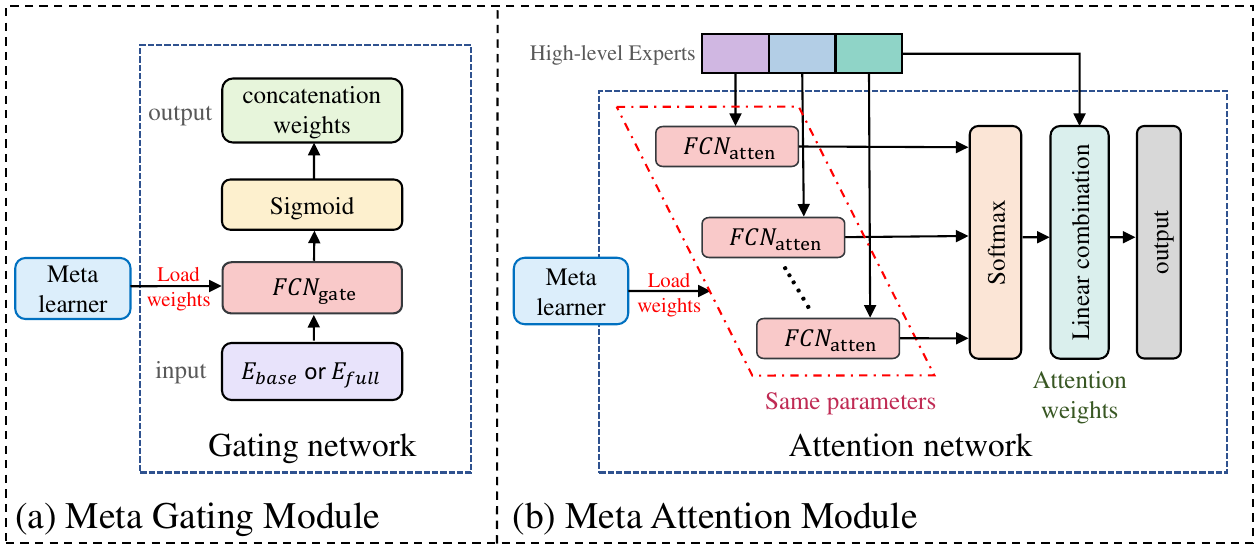} 
    \vspace{-0.56cm}
    \caption{\small Meta Module} 
    \label{fig:meta module}      
    \vspace{-10pt}
\end{figure}
\paragraph{Meta Gating Module}


Building upon the features extracted by the heterogeneous experts, we design a Meta Gating Module to fuse the shared knowledge and specialized knowledge via weighted concatenation. As illustrated in Fig.~\ref{fig:meta module} (a), this module consists of a meta learner and a gating network, where the gating network parameters are directly generated by the meta learner based on domain information.

The gating module dynamically generates a set of scalar weight coefficients conditioned on the context of each order, adaptively adjusting the importance of the $K$ ($K=3$) shared experts and one branch-specific expert. Let $\mathcal{H}_{shared} = \{h_{sh}^{(1)}, h_{sh}^{(2)}, \dots, h_{sh}^{(K)}\}$ denote the set of outputs from the $K$ shared experts. For each branch, we design an independent gating network $\text{Meta}_{\mathcal{G}}$ that produces a $(K+1)$-dimensional scalar weight vector $\mathbf{g} = [g_{sh}^{(1)}, \dots, g_{sh}^{(K)}, g_{spec}]$.

For the Generalization Branch, the gating network $\text{Meta}_{\mathcal{G}}^{gen}$ takes only the basic feature representation $E_{base}$ as input and generates dynamic weights tailored for generalization branch: \begin{equation}\mathbf{g}^{gen} = \text{Sigmoid}(\text{MLP}_{gen}(E_{base}, \Theta_{gate})) \in \mathbb{R}^{K+1}.\end{equation} The resulting $K+1$ scalar weights are then multiplied with the outputs of the corresponding $K$ shared experts and the branch-specific generalization expert, followed by concatenation:\begin{equation}\mathbf{O}_{gen} = \text{Concat}\left( g^{gen}_{spec} \cdot h_{gen}, \quad \{ g^{gen}_{k} \cdot h_{sh}^{(k)} \}_{k=1}^K \right).\end{equation}\noindent For the Source Branch, the gating network $\text{Meta}_{\mathcal{G}}^{src}$ takes the full feature representation $E_{full}$ as input to produces dynamic weights:
\begin{equation}\mathbf{g}^{src} = \text{Sigmoid}(\text{MLP}_{src}(E_{full}, \Theta_{gate})) \in \mathbb{R}^{K+1}\end{equation}
The final output is the weighted concatenation of the $K$ shared experts and the source-specific expert:
\begin{equation}\mathbf{O}_{src} = \text{Concat}\left( g^{src}_{spec} \cdot h_{src}, \quad \{ g^{src}_{k} \cdot h_{sh}^{(k)} \}_{k=1}^K \right)\end{equation}

This design enables fine-grained instance-level adaptation through the context-aware gating. For example, in orders associated with complex traffic conditions, the gating network may automatically increase the weight of the Sequence View Expert ($g_{sh}^{tmp}$) while down-weighting other views.

\paragraph{Meta Attention Module}

After weighted concatenation over the outputs of the heterogeneous experts, we obtain a fused intermediate representation $\mathbf{O}_{branch}$ (i.e., $\mathbf{O}{gen}$ or $\mathbf{O}_{src}$). On top of this, we introduce a second group of expert networks, referred to as High-level Semantic Experts, implemented as $M$ parallel MLPs to capture deeper semantic abstractions: \begin{equation}H_m = \text{MLP}_m(\mathbf{O}), \quad m \in \{1, \dots, M\}\end{equation}where $H_m$ denotes the deep semantic representation extracted by the $m$-th high-level expert. The $M$ MLP experts share the same architecture but maintain independent parameters, and expert groups for the two branches are instantiated separately.

Next, we propose a Meta Attention Module to dynamically aggregate the outputs of the $M$ high-level experts, accounting for domain differences when computing their attention scores. As illustrated in Fig.~\ref{fig:meta module} (b), this module consists of a meta learner and an attention network $\text{Meta}_{Atten}(\cdot)$, whose network parameters $\Theta_{att}$ are generated from the meta learner based on domain information.

For each high-level expert output $H_m$, the attention score $s_m$ is computed as $s_m = \text{Meta}_{Atten}(H_m; \Theta_{att})$. The scores are then normalized via softmax to obtain attention weight vector $\mathbf{a} = [a_1, \dots, a_M]$, and the final representation is aggregated as
\begin{equation}a_m = \frac{\exp(s_m)}{\sum_{k=1}^{M} \exp(s_k)},\,\,\,H_{deep} = \sum_{m=1}^{M} a_m \cdot H_m\end{equation}

\paragraph{Meta Head Module}

At the final prediction stage, we further introduce a Meta Head Module equipped with a meta learner to distinguish different domains at the output layer. By adopting a Wide $\&$ Deep strategy, we fuse the shallow feature $\mathbf{O}_{b}$ (Wide) and the deep representation $H_{deep}^{(b)}$ (Deep), serving as the input for the final projection:\begin{equation}H_{final}^{(b)} = \text{Concat}(\mathbf{O}_{b}, H_{deep}^{(b)}), \,\, Output_{b} = \text{MLP}(H_{final}^{(b)}; \Theta_{Head}^{(b)}),\end{equation}where $b \in \{src, gen\}$, and the head network parameters $\Theta_{Head}^{(b)}$ are generated by the meta learner.

\subsubsection{Meta Learner} 
\label{meta learner}
In the proposed architecture, three critical components are augmented with the meta mechanism, whose parameters are dynamically generated: the Gating Weights ($\Theta_{gate}$) for bottom-level feature aggregation, the Attention Weights ($\Theta_{att}$) for high-level expert fusion, the Head Weights ($\Theta_{Head}$) for the final prediction layer.

\paragraph{Decomposed Meta Mechanism}
Traditional parameter-generating meta-learning typically rely solely on static domain IDs or domain attributes to generate model parameters, producing identical parameters for all samples within the same domain~\cite{zhang2022leaving}. In practice, such coarse-grained parameter generation often leads to underfitting in complex scenarios, especially when only a small number of source domains are available~\cite{ yan2022apg,volk2023example}.

To address this, we decompose the generation of target parameters $\Theta_{target} \in \{\Theta_{gate}, \Theta_{att}, \Theta_{head}\}$ into two components as shown in Fig.~\ref{fig:overview} (B): a domain-attribute-driven domain base and a context-driven instance correction. Formally, for any target parameter matrix $\Theta_{target}$, the generation process is defined as:\begin{equation}\Theta_{target}(x, d_{attr}) = \Theta_{base}(d_{attr}) + \Delta\Theta(x, d_{attr})\end{equation}Here, $d_{attr}$ denotes the domain attribute representation, which consists of dense domain semantic vectors and explicit operational features derived from proprietary surveys. Detailed construction procedures are provided in Section~\ref{sec:Model Implement} and Appendix~\ref{sec: Domain Semantic Knowledge}. The representation $d_{attr}$ is pre-computed and remains fixed during training.

The base term $\Theta_{base}$ captures inter-domain differences and latent correlations, ensuring that domains with similar physical and operational attributes (even if geographically distant) are mapped to similar predictive patterns. For unseen cold-start target domains, the model can generate appropriate parameter based on their domain attributes $d_{attr}$, thereby achieving domain generalization.

The correction term $\Delta\Theta$ captures dynamic variations within a domain, enabling intra-domain adaptation. Even within the same city, delivery patterns can fluctuate significantly due to weather changes or unexpected events. The correction term performs instance-level fine-tuning of the base parameters conditioned on the input features of each order, allowing the model to better align with contextual uncertainty. This dynamic adjustment can allow UME to flexibly adapt its behavior to specific input patterns, especially enhancing robustness in cold-start domain.

\paragraph{Hypernet Architecture Implementation}

For three Meta Modules, each corresponding meta learner is composed of three distinct hypernetworks. Specifically, we first construct a shared base hypernetwork $\mathcal{H}_{base}$ that is shared across two branches. It takes only the domain attribute representation $d_{attr}$ as input: $\Theta_{base} = \mathcal{H}_{base}(d_{attr})$. This hypernetwork ensures that data within the same domain share a stable and unified parameter base. It maintains two separate output heads for generalization branch and source branch, respectively.


Next, considering the difference in available input information between two branches, we design different correction hypernetworks for each branch. For source branch, the correction term is generated using the full context $z_{src} = \text{Concat}(E_{full}, d_{attr})$: $\Delta\Theta_{src} = \mathcal{H}_{res}^{src}(z_{src})$. For generalization branch, the correction term is generated using only the basic context $z_{gen} = \text{Concat}(E_{base}, d_{attr})$: $\Delta\Theta_{gen} = \mathcal{H}_{res}^{gen}(z_{gen})$.

\subsection{Model Optimization and Inference}

\subsubsection{Model Optimization}

As illustrated in Fig.~\ref{fig:ETA}, ETA prediction is influenced by numerous uncontrollable stochastic factors—such as sudden congestion and merchant preparation delays—resulting in significant aleatoric uncertainty~\cite{o2016uncertainty}. Traditional point estimation methods typically assume homoscedastic Gaussian errors. However, this assumption is inconsistent with real-world ETA data. First, ETA data follow a typical right-skewed long-tail distribution. Second, the predictive uncertainty is heteroscedastic, changing with contextual factors such as weather conditions~\cite{wang2018will}. To accurately model this complex stochasticity, we adopt the Gamma distribution as the predictive target distribution.

We design a composite loss function that considers not only predictive accuracy but also the rationality of the predicted distribution. For any branch $b \in \{src, gen\}$, let $\hat{\alpha}_{b}$ and $\hat{\theta}_{b}$ denote the shape and scale parameters of the predicted Gamma distribution, respectively, and $\hat{\mu}_{b} = \hat{\alpha}_{b}\hat{\theta}_{b}$ denotes the predicted expectation. The composite loss $\mathcal{L}_{comp}^{b}$ consists of a Negative Log-Likelihood (NLL) term and a mean-based MSE term:
\begin{equation}
    \mathcal{L}_{comp}^{(b)} = \mathcal{L}_{NLL}(y, \hat{\alpha}_{b}, \hat{\theta}_{b}) + \lambda_{mse} \mathcal{L}_{MSE}(y, \hat{\mu}_{b}),
    \label{comploss}
\end{equation}
\noindent where
\begin{equation}
    \mathcal{L}_{NLL} = -\sum_{i} \left[ (\hat{\alpha}_i - 1)\log y_i - \frac{y_i}{\hat{\theta}_i} - \hat{\alpha}_i \log \hat{\theta}_i - \log \Gamma(\hat{\alpha}_i) \right],
\end{equation}
\begin{equation}
    \mathcal{L}_{MSE} = \sum_{i} (y_i - \hat{\alpha}_i \hat{\theta}_i)^2.
\end{equation}

Although the Generalization Branch is supervised by $\mathcal{L}_{comp}^{gen}$, structural feature missingness makes it difficult to achieve accurate inference using only basic features. To enable it to approximate the predictive capability of the full-feature model, we further introduce a knowledge distillation loss $\mathcal{L}_{KD}$. We employ the Kullback–Leibler (KL) divergence to encourage the predictive distribution of the Generalization Branch (Student), denoted as $P_{gen}$, to approximate the predictive distribution of the Source Branch (Teacher), denoted as $P_{src}$. To preserve the purity of the teacher’s knowledge, we apply a stop-gradient strategy when computing $\mathcal{L}_{KD}$:
\begin{equation}
    \mathcal{L}_{KD} = D_{KL}(\text{stopgrad}(P_{src}) \parallel P_{gen})
    \label{KDloss}
\end{equation}Here, the $\text{stopgrad}(\cdot)$ operator completely blocks the gradient flow from $\mathcal{L}_{KD}$ back to source branch, ensuring $\mathcal{L}_{KD}$ only updates the student parameters.

Finally, the entire framework is jointly trained on the source domain dataset, and the overall loss function is defined as:
\begin{equation}
    \mathcal{L}_{total} = \sum_{(x, x^*, y) \in \mathcal{D}_{train}} \left( \,\underbrace{\mathcal{L}_{comp}^{src}}_{\text{teacher loss}} + \underbrace{\mathcal{L}_{comp}^{gen}}_{\text{student loss}} + \beta \cdot \underbrace{\mathcal{L}_{KD}}_{\text{distillation}} \,\right)
    \label{totalloss}
\end{equation}


\noindent In summary, we refer to the proposed framework as \textbf{UME} (\textbf{U}nified \textbf{M}eta-Generalization Framework for Cross-Domain \textbf{E}TA).

\subsubsection{Model Inference}

During inference on cold-start domains, only the Generalization Branch is activated. Since offline statistical features are unavailable, the model relies exclusively on basic online real-time features for prediction. The Source Branch is disabled in this scenario and does not participate in the forward computation.

During inference on mature source domains, the output of the Source Branch—leveraging full feature information—is used as the final prediction.

\subsection{Model Implement}
\label{sec:Model Implement}

For the domain attribute representation $d_{attr}$, we construct structured textual descriptions for each source and target domain, including information such as city scale, traffic patterns, delivery vehicle types, and climate conditions (see Appendix~\ref{sec: Domain Semantic Knowledge} for details). We employ BGE-Small-en-v1.5, a state-of-the-art pre-trained embedding model~\cite{xiao2024c}, as the feature extractor to encode these descriptions into fixed 384-dimensional semantic vectors. The encoding process is performed offline, and the resulting embeddings remain frozen during training. Subsequently, we concatenate the operational features derived from proprietary internal city-level surveys, forming the final domain attribute representation $d_{attr}$. 

Model optimization is performed using the Adam optimizer. For the distillation loss weight $\beta$ in Eq.~\ref{totalloss}, we adopt a step-wise warm-up strategy to prevent the Generalization Branch (Student) from being misled by low-quality teacher signals during the early training stage when the Source Branch (teacher) has not yet converged. We set $\beta = 0$ during the first 3000 training steps, allowing the student branch to focus on learning fundamental patterns directly from label. Once the teacher branch becomes stable, $\beta$ is restored to enable knowledge distillation. See Appendix~\ref{sec:Sensitivity Analysis} for the sensitivity analysis on the distillation weight and warm-up steps.

\section{Experimental Studies}

\subsection{Experimental Setup}
In our experimental setting, each city is defined as a distinct domain. Both domestic expansion and international market entry are formulated as domain generalization problems, where newly launched cities are treated as unseen target domains.
\subsubsection{\textbf{Datasets}}
We conduct experiments on two large-scale industrial food delivery datasets from the Keeta platform (Meituan’s overseas food delivery service), each containing 63 consecutive days of order logs. To comprehensively evaluate the cross-domain generalization capability of UME, we design two hierarchical settings based on the geographical span between cold-start target cities and source cities.

\textit{Intra-country Generalization Experiments.} This setting focuses on cold-start scenarios where new cities are launched within countries that already have mature operations. The dataset covers 18 cities from: Hong Kong (HK) and Saudi Arabia (SA).

\begin{itemize}[leftmargin=*, noitemsep, topsep=2pt ]
    \item \textbf{Training Set:} Comprises 60 days of order data (Day 1--Day 60) from  Hong Kong and seven cities in Saudi Arabia.
    \item \textbf{Source Testing:} Day 61--63 data from the same training cities.
    
    \item \textbf{Target Testing:} Day 61--63 data from 10 newly launched cities in Saudi Arabia.
\end{itemize}

\textit{Cross-country Generalization Experiments.} This setting evaluates generalization to entirely new markets across different countries or even continents (e.g., leveraging data from Asian countries to predict ETA in newly launched South American markets). The dataset spans six countries: Hong Kong (China), Saudi Arabia, Qatar, Kuwait, the United Arab Emirates (UAE), and Brazil, totaling 23 cities.
\begin{itemize}[leftmargin=*, noitemsep, topsep=2pt ]
    \item \textbf{Training Set:} Day 1--60 data from all cities in five mature markets (Hong Kong, Saudi Arabia, Qatar, Kuwait, and UAE).
   
    \item \textbf{Source Testing:} Day 61--63 data from the same training cities.
    \item \textbf{Target Testing:} Day 61--63 data from Brazil during the initial launch of Santos and São Paulo.
   
\end{itemize}

\begin{figure}[t]
    \centering
    \includegraphics[width=\linewidth]{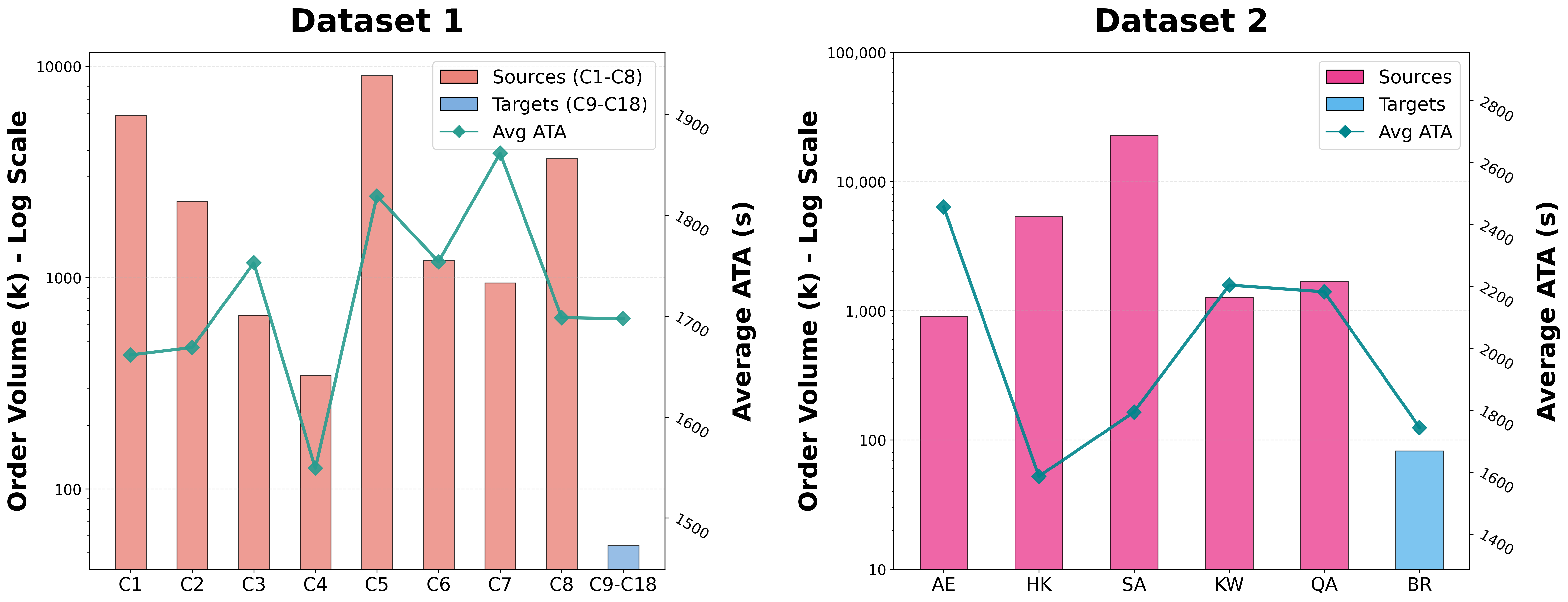}
    \caption{\small Dataset statistics. The bar charts (log-scale) illustrate order volume in source and target domains, while the line plots show the corresponding average ATA.}
    \label{fig:dataset_statistics}
    \vspace{-5pt}
\end{figure}

To illustrate the scale and distribution characteristics of these datasets, Fig.~\ref{fig:dataset_statistics} presents the data statistics for both experimental settings, including the sample size (log-transformed) and average ATA (Actual Time of Arrival) across different domains. In Fig.~\ref{fig:dataset_statistics}(a), the 10 unseen target cities are aggregated into a single group, while Fig.~\ref{fig:dataset_statistics}(b) reports statistics at the country level. Tables~\ref{tab:dataset1_horiz} and~\ref{tab:dataset_horizontal} provide the corresponding raw sample sizes.

\begin{table}[t]
    \centering
    \caption{Statistics of Dataset 1. Order Vol.= Order Volume.}
    \vspace{-8pt}
    \label{tab:dataset1_horiz}
    \resizebox{\linewidth}{!}{
        \begin{tabular}{l|cccccccc|c}
            \toprule
            \textbf{Split} & \multicolumn{8}{c|}{\textbf{Trainset}} & \textbf{Testset} \\
            \midrule
            \textbf{City ID} & C1 & C2 & C3 & C4 & C5 & C6 & C7 & C8 & C9--C18 \\
            \textbf{Order Vol.} & 5847K & 2288K & 664K & 344K & 9010K & 1202K & 944K & 3651K & 53K \\
            \bottomrule
        \end{tabular}
    }
    \vspace{-8pt}
\end{table}

\begin{table}[t]
    \centering
    \caption{Statistics of Dataset 2. Order Vol.= Order Volume.}
    \vspace{-8pt}
    \label{tab:dataset_horizontal}
    \resizebox{\linewidth}{!}{%
    \begin{tabular}{l|ccccc|c}
        \toprule
        \textbf{Split} & \multicolumn{5}{c|}{\textbf{Trainset}} & \textbf{Testset} \\
        \midrule
        \textbf{Country} & HK(China) & Saudi Arabia & Qatar & Kuwait & UAE & Brazil \\
        \textbf{Order Vol.} & 5334K & 22667K & 1686K & 1276K & 902K & 82K \\
        \bottomrule
    \end{tabular}%
    }
    \vspace{-8pt}
\end{table}

\subsubsection{\textbf{Baselines}}

We compare UME against three categories of methods: \textbf{(1)} mainstream industrial ETA models OFCT~\cite{zhu2020order} and DoorDash~\cite{doordash2023precision}; \textbf{(2)} representative domain generalization (DG) methods ARM-CML~\cite{zhang2021adaptive} and SimMMDG~\cite{dong2023simmmdg}; \textbf{(3)} recent state-of-the-art multi-domain learning (MDL) approaches STAR~\cite{sheng2021one}, M2M~\cite{zhang2022leaving}, MultiLoRA~\cite{song2024multilora}, and UFIN~\cite{tian2025ufin}. Detailed descriptions of these baselines are provided as follows:
\begin{itemize}[leftmargin=*, noitemsep, topsep=1pt ]
    \item \textbf{OFCT}~\cite{zhu2020order}: A strong baseline ETA prediction model for the checkout page in on-demand food delivery proposed by Alibaba.

    \item \textbf{DoorDash}~\cite{doordash2023precision}: The latest generation ETA prediction model for checkout page deployed at DoorDash.

    \item \textbf{STAR}~\cite{sheng2021one}: a multi-domain model employing a fully-connected network with a star topology. For zero-shot inference, we use the averaged domain-specific embeddings and parameters learned from source domains, as domain IDs are unavailable for unseen target domains.

    \item \textbf{M2M}~\cite{zhang2022leaving}: a meta-learning-based multi-scenario modeling approach. During domain generalization inference, the global average statistics from the source domains are used to compensate for the missing scenario knowledge in the target domain.

    \item \textbf{ARM-CML} \cite{zhang2021adaptive}: A representative domain generalization framework that employs contextual meta-learning to handle domain shift by extracting latent contexts from unlabeled test inputs.

    \item \textbf{SimMMDG} \cite{dong2023simmmdg}: a multi-modal DG framework that employs cross-modal translation to handle feature missingness. We adapt it by treating feature groups as modalities, enabling the reconstruction of missing offline features from online inputs.

    \item \textbf{MultiLoRA}~\cite{song2024multilora}: a parameter-efficient framework that performs multi-domain learning by fusing multi-directional LoRAs. We utilize its auxiliary domain modules to enable zero-shot generalization to unseen domains.

    \item \textbf{UFIN}~\cite{tian2025ufin}: a recently proposed state-of-the-art multi-domain framework that leverages LLMs to learn universal semantic features across diverse domains.

\end{itemize}

For a fair comparison, all methods are adapted to predict Gamma distributions and are optimized with an additional NLL loss. Since both OFCT and DoorDash do not model domain distinctions, they are trained on mixed data from all domains, with all domains sharing the same model parameters. For all models except SimMMDG, missing offline features and other required inputs and parameters during the target domain testing are imputed using global mean values computed from the source domains.

\subsubsection{\textbf{Evaluation Metrics}}


To comprehensively evaluate UME, we assess performance from two perspectives: Predictive Accuracy and Calibration Reliability.

For accuracy, we use \textbf{MAE} and \textbf{CRPS}.
MAE measures the average absolute deviation between the ground truth $y_i$ and the predicted expectation $\hat{y}_i$.
CRPS~\cite{gneiting2007probabilistic} evaluates the overall quality of the predicted distribution:
\begin{equation}
    \mathrm{CRPS}(\widehat{F}, y) = \int_{-\infty}^{\infty} \left( \widehat{F}(z) - \mathbbm{1}\{z \ge y\} \right)^2 \, \mathrm{d}z
\end{equation}\noindent For calibration, we use \textbf{Reliability Diagrams}~\cite{niculescu2005predicting} based on the Probability Integral Transform (PIT) to assess the alignment between predicted probabilities and observed frequencies. For each test sample $i$, the PIT value is defined as:
\begin{equation}
p_i = \widehat{F}_i(y_i;\hat{\alpha}_i,\hat{\theta}_i) = \int_{0}^{y_i} \frac{1}{\Gamma(\hat{\alpha}_i)\hat{\theta}_i^{\hat{\alpha}_i}} t^{\hat{\alpha}_i-1} e^{-\frac{t}{\hat{\theta}_i}} dt
\end{equation}
For an ideally calibrated forecaster, $\{p_i\}$ should follow $U[0,1]$. We evaluate the empirical CDF of the PIT values at 20 equally spaced points $k/20$ ($k=1,\dots,20$) and plot the observed frequency against the predicted probability; a curve close to the diagonal indicates well-calibrated uncertainty estimates.

\subsection{Main Results}

\begin{table*}[t]
\centering
\caption{\small Performance comparison. Note that we aggregate the performance of the target and source domains by country.}
\vspace{-6.6pt}
\label{tab:performance_comparison}
\resizebox{\textwidth}{!}{%
\begin{tabular}{lcccccc:cccccccccccc}
\toprule
\multirow{4}{*}{\textbf{Model}} & \multicolumn{6}{c:}{\textbf{Intra-country Generalization}} & \multicolumn{12}{c}{\textbf{Cross-country Generalization}} \\
\cmidrule(lr){2-7} \cmidrule(lr){8-19}
& \multicolumn{2}{c}{Target Domains} & \multicolumn{4}{c:}{Source Domains} & \multicolumn{2}{c}{Target Domains} & \multicolumn{10}{c}{Source Domains} \\
\cmidrule(lr){2-3} \cmidrule(lr){4-7} \cmidrule(lr){8-9} \cmidrule(lr){10-19}
& \multicolumn{2}{c}{\textbf{Saudi Arabia}} & \multicolumn{2}{c}{\textbf{Saudi Arabia}} & \multicolumn{2}{c:}{\textbf{HK (China)}} & \multicolumn{2}{c}{\textbf{Brazil}} & \multicolumn{2}{c}{\textbf{HK (China)}} & \multicolumn{2}{c}{\textbf{Saudi Arabia}} & \multicolumn{2}{c}{\textbf{Qatar}} & \multicolumn{2}{c}{\textbf{Kuwait}} & \multicolumn{2}{c}{\textbf{UAE}} \\
\cmidrule(lr){2-3} \cmidrule(lr){4-5} \cmidrule(lr){6-7} \cmidrule(lr){8-9} \cmidrule(lr){10-11} \cmidrule(lr){12-13} \cmidrule(lr){14-15} \cmidrule(lr){16-17} \cmidrule(lr){18-19}
& CRPS & MAE & CRPS & MAE & CRPS & MAE & CRPS & MAE & CRPS & MAE & CRPS & MAE & CRPS & MAE & CRPS & MAE & CRPS & MAE \\
\midrule
OFCT & 4.982 & 0.00\% & 3.851 & 0.00\% & 5.204 & 0.00\% & 5.912 & 0.00\% & 5.183 & 0.00\% & 3.954 & 0.00\% & 3.957 & 0.00\% & 3.822 & 0.00\% & 4.274 & 0.00\% \\
DoorDash & 4.827 & -2.14\% & 3.764 & -1.85\% & 5.182 & +0.65\% & 5.853 & -1.57\% & 5.152 & -1.23\% & 3.872 & -1.48\% & 3.813 & -1.12\% & 3.802 & -1.28\% & 4.351 & +1.17\% \\
STAR & 4.886 & -1.57\% & 3.523 & -8.21\% & 4.853 & -7.03\% & 5.894 & -0.89\% & 4.717 & -8.84\% & 3.398 & -8.15\% & 3.593 & -7.23\% & 3.458 & -8.02\% & 3.917 & -7.97\% \\
M2M& 4.658 & -4.86\% & 3.524 & -8.18\% & 4.676 & -7.86\% & 5.687 & -3.46\% & \underline{4.603} & \underline{-9.23\%} & 3.347 & -9.01\% & 3.494 & -8.54\% & 3.449 & -8.18\% & \underline{3.742} & \underline{-9.37\%} \\
ARM-CML& 4.581 & -7.97\% & 3.583 & -4.50\% & 4.892 & -4.37\% & \underline{5.054} & \underline{-8.21\%} & 4.993 & -4.39\% & 3.493 & -4.53\% & 3.701 & -4.12\% & 3.746 & -4.48\% & 4.112 & -4.23\% \\
SimMMDG & \underline{4.543} & \underline{-8.67\%} & 3.618 & -4.13\% & 4.924 & -4.06\% & 5.186 & -7.58\% & 5.027 & -4.30\% & 3.524 & -4.37\% & 3.727 & -4.97\% & 3.767 & -4.13\% & 4.208 & -4.08\% \\
MultiLoRA & 4.624 & -5.10\% & \underline{3.486} & \underline{-9.95\%} & \underline{4.617} & \underline{-8.58\%} & 5.319 & -5.84\% & 4.614 & -9.12\% & \underline{3.339} & \underline{-9.36\%} & \underline{3.481} & \underline{-8.67\%} & \underline{3.404} & \underline{-9.04\%} & 3.773 & -8.81\% \\
UFIN   & 4.649 & -4.93\% & 3.506 & -9.36\% & 4.681 & -7.47\% & 5.227 & -6.92\% & 4.708 & -7.63\% & 3.407 & -7.92\% & 3.528 & -6.48\% & 3.528 & -6.72\% & 3.891 & -6.53\% \\
\midrule
\textbf{UME} & \textbf{4.361} & \textbf{-11.85\%} & \textbf{3.190} & \textbf{-11.50\%} & \textbf{4.469} & \textbf{-9.60\%} & \textbf{4.826} & \textbf{-12.43\%} & \textbf{4.414} & \textbf{-11.14\%} & \textbf{3.187} & \textbf{-11.87\%} & \textbf{3.373} & \textbf{-10.92\%} & \textbf{3.328} & \textbf{-12.03\%} & \textbf{3.654} & \textbf{-11.56\%} \\
\bottomrule
\end{tabular}%
}
\end{table*}

For accuracy analysis, Table~\ref{tab:performance_comparison} presents a comprehensive comparison between UME and baseline methods on unseen target domains with structural feature missingness and mature source domains. MAE is reported as the relative percentage reduction compared to OFCT (0.00\%), where more negative values indicate greater improvement. The best results among all baselines are underlined to facilitate comparison. Single-domain models (OFCT, DoorDash) perform the worst due to their inability to model cross-domain discrepancies. Multi-domain approaches such as STAR and M2M achieve competitive results on source domains but struggle on unseen targets, as they rely on parameter averaging or limited domain-level knowledge, which fails to capture new-market heterogeneity and may lead to underfitting. Domain generalization methods (ARM-CML, SimMMDG) achieve reasonable performance on target-domain evaluations, while recent multi-domain methods (MultiLora, UFIN), despite performing well on source domains, remain limited by the absence of explicit DG mechanisms.

None of these methods effectively address structural feature missingness; instead, they rely on global mean imputation, resulting in degraded performance in cold-start settings. Even SimMMDG, which attempts offline feature reconstruction, cannot recover high-value historical information that is inherently absent in new domains. In contrast, by explicitly incorporating a domain generalization strategy and a principled treatment of structural feature missingness, UME consistently achieves significantly superior performance across both source and unseen target domains.

\begin{figure*}[tbp]
    \centering  
    
    \includegraphics[width=1.0\textwidth]{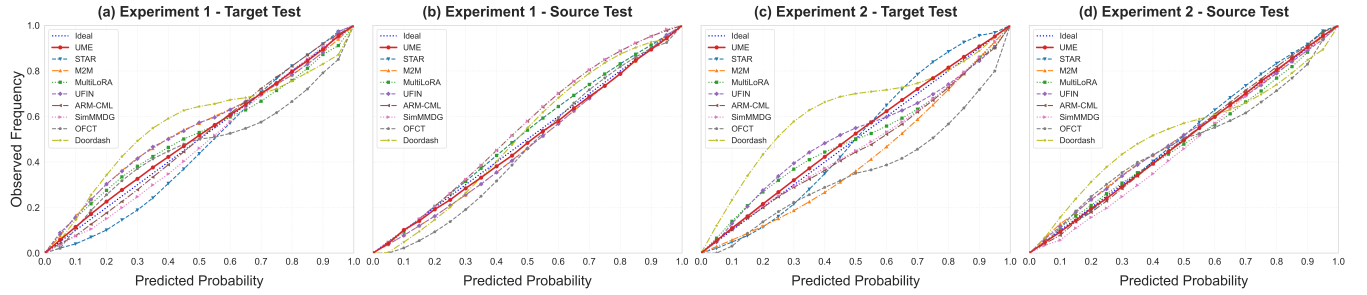} 
    \vspace{-0.65cm}
    \caption{\small Comparison of Reliability Diagrams} 
    \label{fig:overall_reliability}     
    \vspace{-3pt}
\end{figure*}

For calibration analysis, Fig.~\ref{fig:overall_reliability} presents the reliability diagrams. Some models exhibit overconfidence, and others are overly conservative, especially in cold-start target domains. However, UME (red line) closely aligns with the diagonal “Ideal” line across all settings, demonstrating superior calibration and more reliable uncertainty estimation than all baselines.

\subsection{Ablation Studies}

To validate the effectiveness of the key components in UME, we conduct comprehensive ablation studies on the cross-country dataset. The full model is compared with four variants:\begin{itemize}[leftmargin=*, noitemsep, topsep=3pt ]
\item \textit{w/o Unified}: This variant represents an independent modeling scheme. We separately train two models: a mature-operation model using full features (evaluated on sources) and a cold-start model using only basic features (evaluated on targets).

\item \textit{w/o Meta-Mechanism}: This variant removes the meta-learning mechanism. The parameters of the gating module, attention module, and output heads are replaced with static learnable parameters shared across all domains.

\item \textit{w/o Dynamic Correction}: This variant retains only the domain-level base parameter generation driven by domain attributes $d_{attr}$, while removing the instance-level correction. In this setting, the target parameters degenerate to $\Theta_{target} = \Theta_{base}(d_{attr}) $.

\item \textit{w/o Distillation}: This variant removes the distillation strategy by excluding the distillation loss $\mathcal{L}_{KD}$.

\end{itemize}As shown in Table~\ref{tab:ablation}, UME achieves the best performance across all metrics. Removing the meta mechanism leads to the largest degradation, demonstrating that meta learning is essential for handling domain shifts. To further investigate the individual contributions of three modules, we provide a fine-grained ablation analysis in Appendix~\ref{sec:Additional Ablation Studies}. The \textit{w/o Dynamic Correction} variant performs second worst, showing that only domain-level modulation without instance-level correction ($\Delta\Theta$) cannot adequately capture diverse domain patterns. Although \textit{w/o Unified} improves over the two variants above, it remains inferior to the unified design. Moreover, source-domain evaluations show that UME also outperforms independent modeling in mature domains, indicating that unified modeling enables mutual enhancement between cold-start and mature-operation prediction. Finally, removing distillation degrades performance in cold-start target domains, indicating that knowledge transfer helps the Generalization Branch better approximate the teacher under feature-missing conditions.

\begin{table}[htbp]
\vspace{-0.05cm}
  \centering
  \caption{Ablation studies of UME.}
  \vspace{-8pt}
  \label{tab:ablation}
  \small
  \begin{tabular}{l|cc|cc}
    \toprule
    \multirow{2}{*}{Method} & \multicolumn{2}{c|}{Cold Start Targets} & \multicolumn{2}{c}{Mature Sources} \\
    & MAE & CRPS & MAE & CRPS \\
    \midrule
    w/o Unified & 6.16\% & 5.21 & 2.15\% & 3.67 \\
    w/o Meta-Mech & 14.28\% & 6.67 & 9.57\% & 4.73 \\
    w/o Corr-term & 6.56\% & 5.34 & 3.22\% & 3.83 \\
    w/o Distill & 5.19\% & 5.10 & 0.91\% & 3.60 \\
    \textbf{UME (Full)} & \textbf{0\%} & \textbf{4.83} & \textbf{0\%} & \textbf{3.58} \\
    \bottomrule
  \end{tabular}
  \vspace{-2.5pt}
\end{table}

\subsection{Online A/B Test}

UME has been successfully deployed in the production environment of Meituan-Keeta platform. To evaluate its post-launch performance, we conducted testing when City X—the first city launched in a new country market on the Keeta platform—went live. We conducted a 20-day online A/B test against the production baseline model by splitting traffic in a 50/50 ratio. In production pipeline, both UME and the baseline are retrained every three days using the most recent 60-day sample set. Data collected from the new city was used exclusively to update the Generalization Branch. Beyond MAE, we introduced the Bad Case Rate (BCR) as a key business metric, defined as the proportion of orders with severe delivery delays.

Table~\ref{tab:online_ab_results} reports the relative improvements of UME over the baseline. In cold-start domain, MAE and BCR are reduced by 15.43\% and 18.72\%, respectively. In mature domains, UME further achieves a 6.31\% reduction in MAE and a 17.24\% reduction in BCR. In addition, during the online testing, the number of customer complaints related to inaccurate ETA drops significantly by 58.92\%. These results demonstrate that UME not only enhances predictive accuracy but also improves service reliability and customer satisfaction.
\begin{table}[htbp]
  \centering
  \caption{Online A/B Testing Results}
  \vspace{-8pt}
  \label{tab:online_ab_results}
  \small
  \begin{tabular}{l|cc|cc}
    \toprule
    \multirow{2}{*}{Model} & \multicolumn{2}{c|}{Cold Start Targets} & \multicolumn{2}{c}{Mature Sources} \\
    & MAE  & BCR  & MAE  & BCR  \\
    \midrule
    UME& $-15.43\%$ & $-18.72\%$ & $-6.31\%$ & $-17.24\%$ \\
    \bottomrule
  \end{tabular}
  \vspace{-9.5pt}
\end{table}

\subsection{Discussion}
\label{sec:Discussion}
This section further discusses the scalability of the UME framework in industrial deployment. Although our experiments primarily focus on the highly challenging zero-shot cold-start scenario during the early stage of new city launch, UME is inherently scalable to accommodate the full operational lifecycle of every domain. In terms of model evolution, as data gradually accumulate in the target domain (e.g., a few days after launch), UME can transition from a zero-shot setting to a few-shot setting. The newly collected cold-start are used exclusively to update the Generalization Branch, enabling the meta learner to refine its parameters without modifying the overall architecture.

From the perspective of industrial reliability, the value of the Generalization Branch extends beyond cold-start stage. It also serves as a robust fallback mechanism in mature domains. When offline features experience severe distribution shifts due to rare events—such as Ramadan in Saudi Arabia, major holidays, or large-scale public events—leading to the breakdown of historical statistics, the model can rely on the Generalization Branch to maintain stable predictions. This design mitigates the failure risk that may arise in single-branch models overly dependent on offline features.

\section{Conclusion}

This paper addresses the challenges of domain shift and zero-shot generalization in cross-domain ETA prediction under structural feature missingness. We propose UME, a unified framework that integrates a dual branch architecture and meta learning to support prediction in both cold-start and mature-operation domains. UME has been deployed on the Meituan-Keeta. Extensive offline experiments and online A/B tests demonstrate that UME achieves the best accuracy and calibration performance. By eliminating the dependency for separate models across different operational stages, UME provides a scalable and low-maintenance solution for global instant delivery platforms. Future work will fine-tune the text embedding model into a specialized domain-knowledge extractor to generate domain representations tailored for ETA prediction.

\bibliographystyle{ACM-Reference-Format}

\balance
\bibliography{sample-base}

\appendix

\section{Domain Semantic Knowledge}
\label{sec: Domain Semantic Knowledge}

We construct a unified prompt template to standardize the textual profile of each domain. This standardization ensures that the resulting embeddings accurately reflect the heterogeneity in delivery environments across different regions. We encode structured domain descriptions into semantic vectors using BGE-Small-en-v1.5, a lightweight yet powerful embedding model that captures fine-grained city semantics while maintaining a compact 384-dimensional space to prevent hypernetwork parameter explosion. The template is defined as follows:

"A food delivery operational environment in [City Name]. The urban layout is characterized by [Urban Scale \& Road Density]. The traffic condition is generally [Traffic Pattern]. The dominant delivery vehicle used by couriers is [Vehicle Type]. The local climate is [Climate Type], which impacts outdoor delivery conditions."

Concrete examples in Table.~\ref{semanticemb} demonstrates how different source and target domains are instantiated using this protocol.

\begin{table*}[htbp] 
  \centering
  \caption{Examples of Structured Domain Descriptions used for Semantic Embedding Generation.}
  \vspace{-5pt}
  \label{tab:domain_descriptions}
  \renewcommand{\arraystretch}{1.2} 
  \begin{tabularx}{\textwidth}{@{}p{3.2cm} X p{4cm}@{}} 
    \toprule
    \textbf{Domain Type} & \textbf{City Profile (Input to PLM)} & \textbf{Physical Implications} \\
    \midrule
    
    \textbf{Source Domain A} \newline \textit{(e.g., Mega-city)} & 
    ``A food delivery operational environment in \textbf{City A}. The urban layout is characterized by \textbf{extremely high-density road networks with complex overpasses}. The traffic condition is \textbf{highly congested during peaks}. The dominant delivery vehicle is \textbf{E-bikes}. The local climate is \textbf{temperate with seasonal rain}.'' & 
    High uncertainty; flexible routing via e-scooters; moderate weather impact. \\
    \midrule
    
    \textbf{Source Domain B} \newline \textit{(e.g., Small town)} & 
    ``A food delivery operational environment in \textbf{City B}. The urban layout is characterized by \textbf{small-scale grids and low density}. The traffic condition is \textbf{stable and free-flowing}. The dominant delivery vehicle is \textbf{electric scooters}. The local climate is \textbf{mild and dry}.'' & 
    Low uncertainty; high average speed; minimal weather impact. \\
    \midrule
    
    \textbf{Target Domain C} \newline \textit{(e.g., High-density Metropolis)} & 
    ``A food delivery operational environment in \textbf{City C}. The urban layout is characterized by \textbf{dense skyscrapers and narrow alleys}. The traffic condition is \textbf{complex with limited vehicle access}. The dominant delivery vehicle is \textbf{walking or cycling}. The local climate is \textbf{humid subtropical}.'' & 
    Very short distances but long last-mile delivery time; reliance on human physical stamina. \\
    
    \bottomrule
  \end{tabularx}
  \label{semanticemb}
\end{table*}

\section{Additional Ablation Studies}
\label{sec:Additional Ablation Studies}

\begin{figure}[t]
  \centering
  \includegraphics[width=0.9\linewidth]{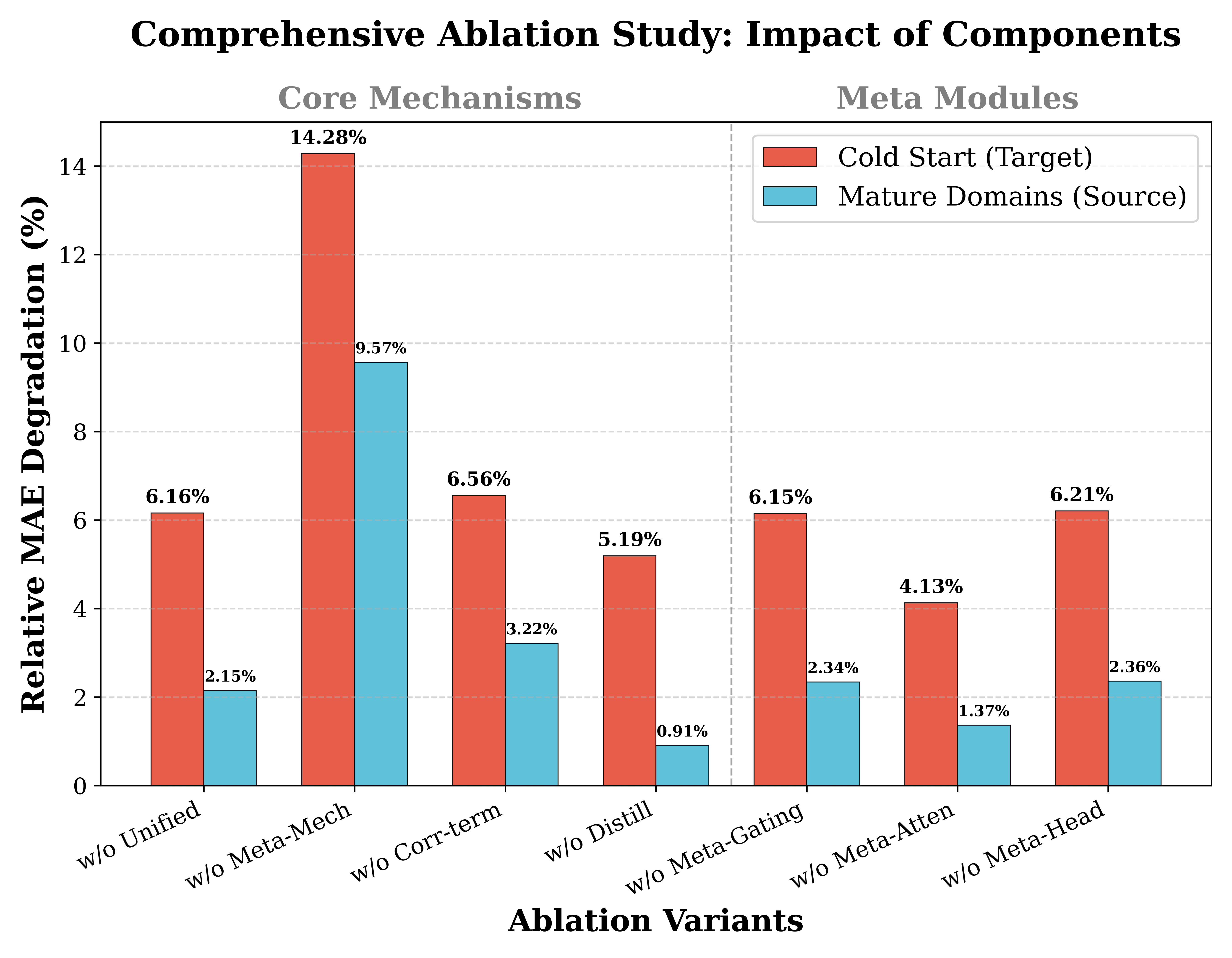}
  \vspace{-0.25cm}
  \caption{\small \textbf{Ablation studies of UME}.The y-axis shows the relative MAE degradation compared to the full UME model (lower is better).}
  \label{fig:ablation_fig}
  \vspace{-6pt}
\end{figure}

\begin{figure*}[t] 
    \centering
    
    \includegraphics[width=\textwidth]{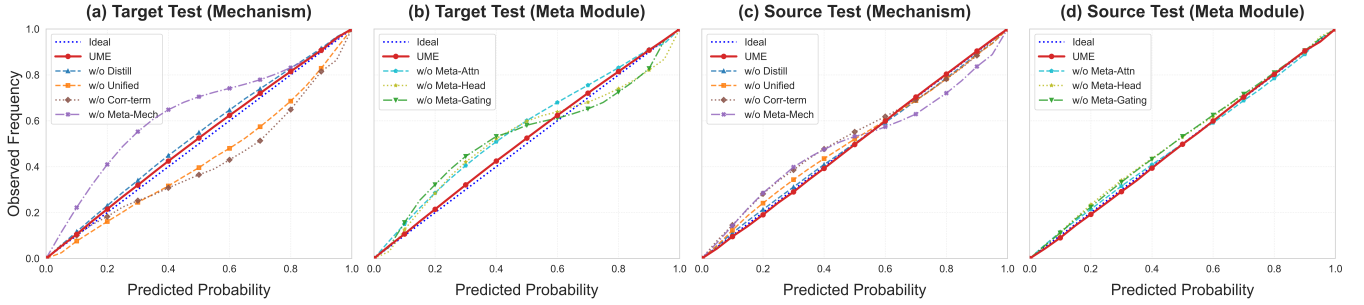} 
    \vspace{-0.5cm}
    \caption{\small \textbf{Reliability diagrams for the ablation study.} \textbf{(a)} and \textbf{(b)} show results on the cold start target Domain.
    \textbf{(c)} and \textbf{(d)} show the corresponding results on the mature source domains. The diagonal line represents perfect calibration. UME consistently exhibits lower calibration error (closer to the diagonal) compared to all baselines.}
    
    \label{fig:ablation_reliability}
\end{figure*}

To further investigate the individual contributions of the three core meta components (Meta Gating, Meta Attention, and Meta Head) in UME, we conduct fine-grained ablation studies. Specifically, while keeping all other components unchanged, We remove one meta component at a time and replace it with a static-parameter network shared across all domains. The resulting variants are denoted as \textit{w/o Meta Gating}, \textit{w/o Meta Attention}, and \textit{w/o Meta Head}. The results are reported in Table~\ref{tab:add_ablation}, Fig.~\ref{fig:ablation_fig} and Fig.~\ref{fig:ablation_reliability}, from which several observations can be drawn.

Removing any meta component leads to consistent degradation in both MAE and CRPS, demonstrating that all three modules play indispensable roles in modeling domain discrepancies across different levels. The largest performance drops occur in \textit{w/o Meta Gating} and \textit{w/o Meta Head} in the cold-start evaluation, highlighting the critical role of domain-aware feature reweighting and output modeling in cross-domain generalization. In contrast, removing Meta Attention leads to a smaller degradation, likely because high-level experts already capture relatively robust and transferable representations. Furthermore, regarding probabilistic calibration, Fig.~\ref{fig:ablation_reliability} visually confirms that removing these meta modules leads to significant deviations from the ideal reliability curve, especially in cold-start target domains. In addition, Fig.~\ref{fig:ablation_fig} show an intuitive comparison of MAE performance variations across all ablation variants.

\begin{table}[htbp]
  \centering
  \caption{Ablation studies on individual Meta Modules.}
  \vspace{-8pt}
  \label{tab:add_ablation}
  \small
  \begin{tabular}{l|cc|cc}
    \toprule
    \multirow{2}{*}{Method} & \multicolumn{2}{c|}{Cold Start Domains} & \multicolumn{2}{c}{Mature Domains} \\
    & MAE & CRPS & MAE & CRPS \\
    \midrule
    w/o Meta-Gating & 6.15\% & 5.19 & 2.34\% & 3.69 \\
    w/o Meta-Atten & 4.13\% & 4.90 & 1.37\% & 3.63 \\
    w/o Meta-Head & 6.21\% & 5.26 & 2.36\% & 3.74 \\
    \textbf{UME (Full)} & \textbf{0\%} & \textbf{4.83} & \textbf{0\%} & \textbf{3.58} \\
    \bottomrule
  \end{tabular}
\end{table}

\begin{figure}[H]
    \centering
  
    \includegraphics[width=\linewidth]{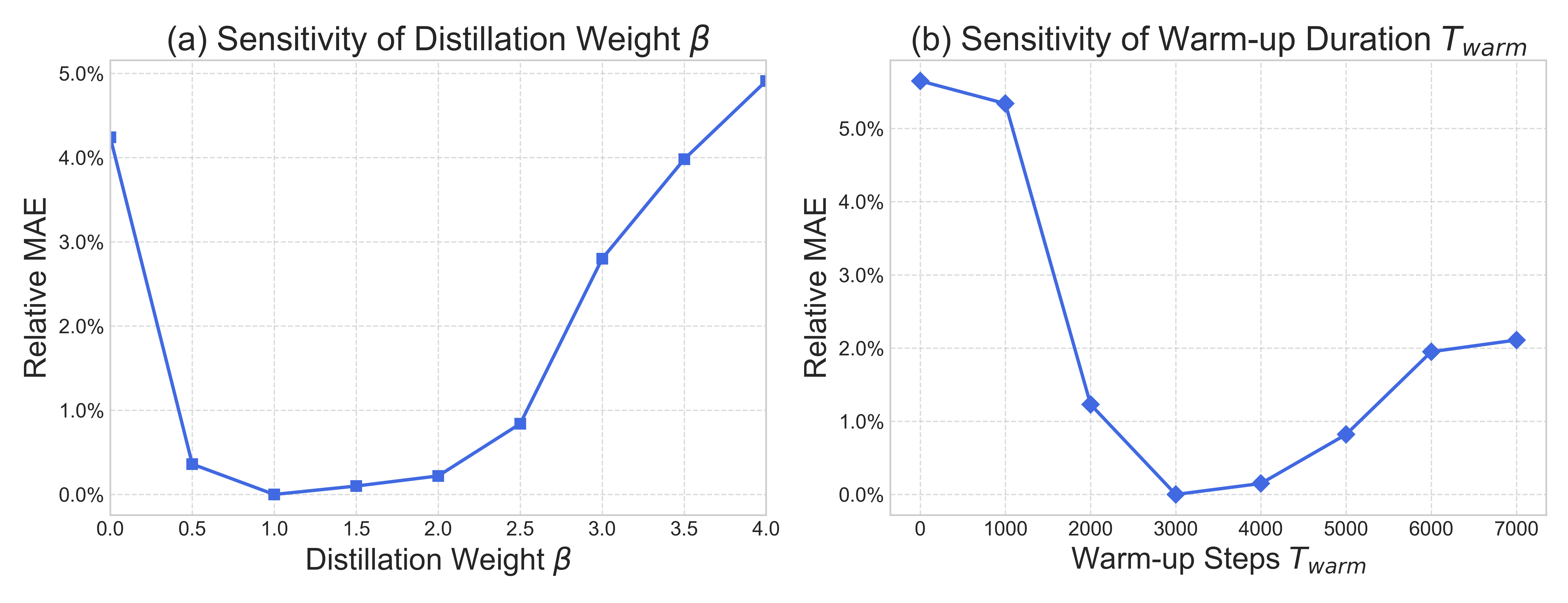}
    \vspace{-5.5pt}
    \caption{\small Sensitivity analysis of hyperparameters: (a) Impact of distillation weight \protect$\beta$, and (b) Impact of warm-up steps $T_{warm}$. The y-axis represents the relative MAE degradation compared to the optimal setting (lower is better).}
    \label{fig:combined_sensitivity}
\end{figure}

\begin{figure}[H]
    \centering
    \includegraphics[width=0.8\linewidth]{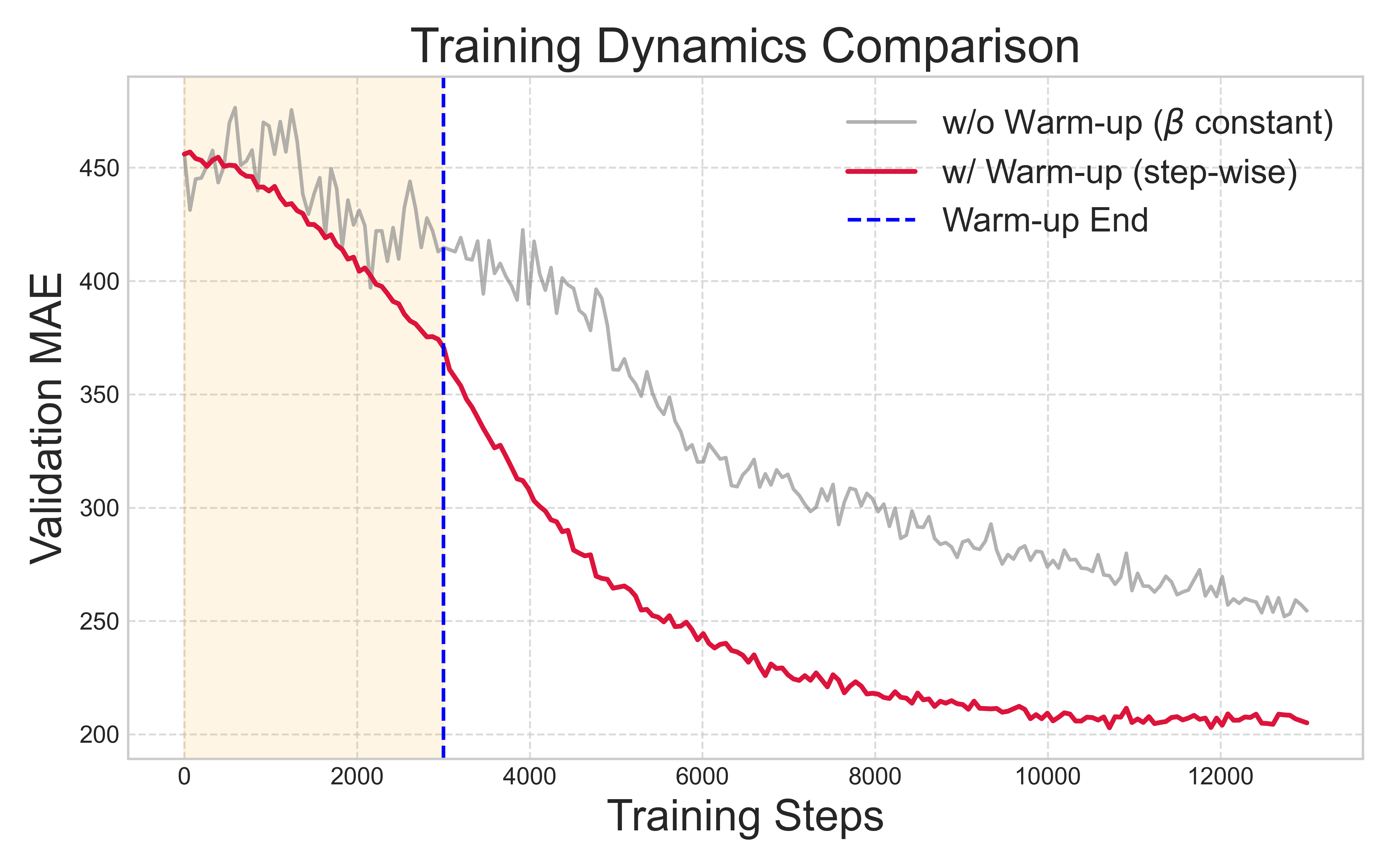} 
    \vspace{-5.5pt}
    \caption{\small Training dynamics comparison between UME with the step-wise warm-up strategy and a baseline with constant distillation ($T_{warm}=0$). The warm-up strategy ensures stable convergence and leads to a lower final MAE.}
    \label{fig:training_dynamics}
\end{figure}

\section{Sensitivity Analysis}
\label{sec:Sensitivity Analysis}
To evaluate the robustness of the knowledge distillation strategy in UME and examine the necessity of the Warm-up during early training, we conduct a sensitivity analysis on two key hyperparameters on the Cross-country Generalization dataset: the distillation weight $\beta$ and the Warm-up steps $T_{warm}$. For both sensitivity studies, we report the relative MAE of the Generalization Branch on the cold-start test set, measured with respect to the best-performing configuration.

\subsection{Effect of Distillation Weight \texorpdfstring{$\beta$}{beta}}

As shown in Fig.~\ref{fig:combined_sensitivity} (a), we vary $\beta \in \{0, 0.5, 1,1.5,\dots, 4\}$ while fixing $T_{warm}=3000$. The results indicate that model performance remains stable within a wide range of $\beta \in [0.5, 2.0]$, demonstrating that the method is insensitive to the distillation weight and thus suitable for practical deployment. Both excessively small and overly large values of $\beta$ lead to performance degradation, further validating the choice of $\beta=1.0$ in the main experiments.

\subsection{Warm-Up}

As illustrated in Fig.~\ref{fig:combined_sensitivity} (b) ($\beta=1$), when $T_{warm}=0$, the error is substantially higher than that with $T_{warm}=3000$, and even worse than \textit{w/o Distill} in Table.~\ref{tab:ablation}. This suggests that enforcing distillation before the Source Branch has stabilized introduces noisy supervisory signals, resulting in negative transfer.

\subsection{Training Dynamics}

We construct a cold-start validation set and visualize the validation MAE of the Generalization Branch during training (for confidentiality, MAE values are transformed in the figure). As shown in Fig.~\ref{fig:training_dynamics}, without Warm-up, the model exhibits pronounced oscillation and delayed convergence in early stages. In contrast, with Warm-up, the error drops sharply once the Teacher is introduced (at 3000 steps), leading to more stable training dynamics and faster convergence.

\end{document}